# END-TO-END SPOKEN LANGUAGE UNDERSTANDING USING TRANSFORMER NETWORKS AND SELF-SUPERVISED PRE-TRAINED FEATURES

*Edmilson Morais, Hong-Kwang J. Kuo, Samuel Thomas, Zoltán Tüske and Brian Kingsbury*

IBM Research

## ABSTRACT

Transformer networks and self-supervised pre-training have consistently delivered state-of-art results in the field of natural language processing (NLP); however, their merits in the field of spoken language understanding (SLU) still need further investigation. In this paper we introduce a modular End-to-End (E2E) SLU transformer network based architecture which allows the use of self-supervised pre-trained acoustic features, pre-trained model initialization and multi-task training. Several SLU experiments for predicting intent and entity labels/values using the ATIS dataset are performed. These experiments investigate the interaction of pre-trained model initialization and multi-task training with either traditional filterbank or self-supervised pre-trained acoustic features. Results show not only that self-supervised pre-trained acoustic features outperform filterbank features in almost all the experiments, but also that when these features are used in combination with multi-task training, they almost eliminate the necessity of pre-trained model initialization.

## 1. INTRODUCTION

Spoken Language Understanding (SLU) systems tackle the problem of inferring semantic information from the speech signal. Traditional approaches consist of a pipeline of automatic speech recognition (ASR) and natural language understanding (NLU) modules [2,3]. Given that the NLU model in this framework is usually trained on clean transcriptions, its application on erroneous ASR transcriptions often reduces the final SLU performance. Although this cascaded approach is widely adopted, there is a growing interest for End-to-End (E2E) SLU which combines ASR and NLU into a single model, avoiding not only cumulative ASR and NLU errors, but also preserving rich prosodic information (e.g., speech rate, pitch and intonation) which is inevitably lost after ASR [8,26].

In order to perform E2E SLU experiments several research groups have built models using multiple stacks of RNNs [5,9,12] to encode the entire utterance as a vector which is fed to a fully connected feedforward neural network followed by a softmax or max-pool layer to identify, for example, domain, intent, and entities. These models work as a classifier, treating each unique combination of domain, intent, and entities as an output label. The limitation of this approach is that the combination of such labels may grow exponentially; moreover, the number of domains, intents, and entities is usually not fixed, which limits the usability of such classification based approach.

A natural approach to deal with the variable-length output for E2E SLU is to use sequence-to-sequence (seq2seq) models. In [8], several RNN based seq2seq architectures are proposed for E2E SLU, among which the authors found that multi-task learning is a key factor for model performance improvement. More recently, other RNN based seq2seq models have been proposed by [12], highlighting the importance of model pre-training. The first Transformer based seq2seq model for E2E SLU was introduced in [6]; however, the authors used an architecture which supports neither multi-task learning nor model pre-training.

Self-supervised pre-trained models have consistently delivered state-of-art results in the field of natural language processing (NLP); however, their merits in the field of spoken language understanding (SLU) still need further investigation. In NLP, BERT [17], UniLM [18] and BART [19] have been successfully applied in language inference [20], question answering [21] and summarization [22]. In speech processing APC [1], wav2vec [16], vq-wav2vec [14], wav2vec 2.0 [15], Mockingjay [11], Tera [13], and audio Albert [4] have been applied for tasks, such as phoneme recognition, automatic speech recognition, speaker recognition, sentiment analysis and speech to text translation [10]. However, to the best of our knowledge, none of these self-supervised speech pre-training techniques have been applied to the field of SLU. As far as we know, all the SLU pre-trained models up to now have been trained in a supervised fashion using labeled speech data, e.g., words or phonemes corresponding to audio signals [12]. As a result, the massive unlabeled speech data available nowadays cannot be utilized by these models.

Given this scenario, the two main goals of this paper are: (1) to introduce a modular transformer-based architecture for E2E SLU, built on top of the well-known ESPNet toolkit [7] and (2) to present a sequence of experiments comparing/analyzing the performance of the proposed E2E SLU system under different configurations: traditional acoustic features (filterbank), self-supervised pre-trained acoustic features (wav2vec), pre-trained model initialization, and multi-task training. It is important to emphasize that, despite achieving competitive results, the hyperparameters of our E2E SLU model have not been fully optimized yet, and because of that, state-of-the-art results have not been included as one of the main goals of this paper. The rest of this paper is organized as follows: Section 2 presents the proposed End-to-End SLU framework. The experimental setup is presented in Section 3. Section 4 presents the results. Finally, conclusions and future work are presented in Section 5.

## 2. PROPOSED END-TO-END SLU MODEL

In this paper we formulate the E2E SLU problem as seq2seq (encoder-decoder) mapping from the continuous speech domain to the symbolic semantic representation domain.

Speech input is represented as sequence of feature vectors and the symbolic semantic output as a sequence of Intent labels and Entity labels/values (I&E) in accordance with the IOB format [23]. For example, let's assume that the full word transcription of a given speech input is:

*I would like to find a flight from charlotte to las vegas that makes a stop in saint louis*

Then, the intent and entity output sequence (I&E) will be represented in the IOB format as:

**O-INT-flight** *charlotte* **B-fromloc-city-name** *las* **B-toloc-city-name** *vegas* **I-toloc-city-name** *saint* **B-stoploc-city-name** *louis* **I-stoploc-city-name** **O-INT-flight**

Note that only words that carry useful semantic meaning are kept in the final I&E output: *charlotte, las vegas* and *saint louis*. In addition to that, the intent label **O-INT-flight** is also inserted at the beginning and end of the semantic output sequence.

### 2.1. Multi-task model architecture

The proposed seq2seq E2E SLU model was built using the ESPNet toolkit [7]. As can be seen from Figure 1, our E2E SLU model is trained with a multi-task objective on three separate tasks, each of them modeled as a simple seq2seq model built from Encoder (E) and Decoder (D) blocks:

1. S2I&E: Speech to Semantics (*main task*)
2. S2T: Speech to Text (*auxiliary task*)
3. T2I&E: Text to Semantics (*auxiliary task*)

The S2I&E seq2seq mapping corresponds to the main SLU task. The S2T and T2I&E are auxiliary tasks well related to the main S2I&E task and are designed to encourage a better learning process of the main task [27].

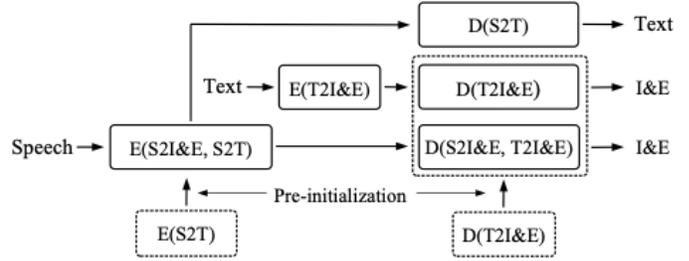

**Figure 1:** Architecture of the proposed E2E SLU model.

It is possible to argue that the S2T task will have a strong correlation with the entity values, which correspond to words in the text transcription. It is also possible to argue that the T2I&E corresponds to a cross-modal auxiliary task that maps word transcriptions to the semantic domain without implicitly deriving information from the speech domain.

During the training stage, the SLU loss function of the overall model will be given by a weighted sum of the loss functions of each task, as described below:

$$SLU\_loss = S2I\&E\_weight \cdot S2I\&E\_loss +$$
$$S2T\_weight \cdot S2T\_loss +$$
$$T2I\&E\_weight \cdot T2I\&E\_loss$$

All the encoders and decoders of our E2E SLU model were built using Transformers networks. The encoders from the S2I&E, S2T and T2I&E tasks have exactly the same configuration. The same is also true for the three decoders corresponding to each task. In addition to that, the encoders from tasks S2I&E and S2T are tied together E(S2I&E, S2T). Similarly, the decoders from tasks S2I&E and T2I&E are also tied together D(S2I&E, T2I&E).

As can be seen from Figure 1, the proposed E2E SLU model also allows pre-initialization of the Encoder (E) and Decoder (D) blocks of the main task using a pre-trained Encoder from an independent S2T system and a pre-trained decoder from another independent T2I&E system, respectively. It is important to emphasize that these independently pre-trained encoder from S2T and decoder from T2I&E systems need to be constructed using the same configurations used in the corresponding encoder and decoder of S2I&E.

### 2.2. Dataset

We used the ATIS dataset compiled in [9], which contains 4976 training and 893 test utterances. The 4976 training utterances comprise ~9.64 hours of audio from 355 speakers. The 893 test utterances comprise ~1.43 hours of audio from 55 speakers. Different from [9], we used the original 16kHz version of ATIS and we augmented the data using only speed perturbation by factors of 0.9, 1.0 and 1.1, expanding the audio training data to ~30 hours. The ATIS dataset was used in this paper for both the overall training of the E2E SLU model and optionally for pre-training the encoder of S2T models and the decoder of T2I&E models that were used to pre-initialize the encoder/decoder of the main S2I&E task.

Two subsets with 960h and 100h of speech from the Librispeech datasets were also used in this paper for pre-training the encoder of S2T models that were used to pre-initialize the encoder of the main task S2I&E of our E2E SLU model.

### 2.3. Evaluation metrics

We measure entity label and value prediction performance with the F1 score presented in [9]. This F1 score requires that both the entity label and entity value must be correct. For example, if the reference is *toloc.city_name:new_york* but the decoded output is *toloc.city_name:york*, then we count both a false negative and a false positive. It is not sufficient that the correct entity label is produced and no "partial credit" is given for part of the entity value (*york*) being recognized.

We measure the intent recognition performance using a simple Intent Error Rate (IER) classification metric.

### 2.4. Acoustic and linguistic encoding

The speech signals were parametrized at a frame rate of 10 ms using both: (1) traditional filterbank (plus pitch information), with 83 coefficients [24] and (2) self-supervised wav2vec [16] with 512 coefficients.

In order to ensure a fair comparison between the feature dimensions of the filterbank and wav2vec features, a linear layer followed by a rectified linear unit was introduced at the input of the encoder modules of the S2T and S2I&E tasks to reduce the input dimension from 512 to 83.

Word transcriptions and semantic representations (I&E) were encoded using byte pair encoding [25]. The semantic representation from ATIS (I&E) was encoded using 512 word pieces. Word transcriptions for Librispeech and ATIS were encoded using 5000 and 512 word pieces, respectively.

## 3. EXPERIMENTAL SETUP

Among the several experiments that can be performed using our E2E SLU model, we have selected experiments to understand the following: (1) importance of model pre-initialization, (2) role of multi-task learning (3) comparison of using traditional acoustic features (log-mel features) with self-supervised features from wav2vec. Results of these experiments are summarized in Table 1 and described in detail below:

In columns 3 and 4 in Table 1, under the term **Pre-Initialization**, we indicate whether either Encoder or Decoder or both Encoder and Decoder of the main task S2I&E have been pre-initialized. For example, in Experiment 1, neither encoder nor decoder have been pre-initialized, in Experiment 5 both encoder and decoder have been pre-initialized using pre-trained encoder and decoder trained using ATIS dataset and in Experiment 9 only the encoder has been pre-initialized using a pre-trained encoder trained using LibriSpeech with 100 hours.

In columns 5 and 6 of Table 1, under the term **Auxiliary Tasks**, we show whether the auxiliary tasks S2T or T2I&E or both S2T and T2I&E were used during the training of the overall E2E SLU model. For example, in Experiment 1 neither S2T nor T2I&E auxiliary task were used during training; in Experiment 7 only S2T task, and in Experiment 10 both S2T and T2I&E were used during training.

In columns 7 and 8 of Table 1, under **Entities (F1 score %)**, entity prediction for both filterbank and wav2vec are presented. Similarly, in columns 9 and 10, under the term **Intent (IER %)** the Intent Error Rate for both filterbank and wav2vec are shown.

In order to facilitate a further analysis of these experiments we grouped them under three main sets described in column 1 of Table 1.

*Set 1* [No pretraining] – In these experiments (1-4), neither the encoder nor the decoder of the main task S2I&E were initialized at all.

*Set 2* [Light pretraining] – Experiments (5-8). In this category encoder and decoder have their components initialized with models trained on the ATIS dataset.

*Set 3* [Enhanced pretraining] – In this set of experiments (9-12), the encoder of the main task S2I&E was either pre-trained on 100 hours of Librispeech (experiments 9-10) or 960 hours of Librispeech (experiments 11-12). Experiments using wav2vec were not done for 960 hours of Librispeech.

### 3.1. Reference models

In order to have baselines to compare with the experiments from Figure 1, we built and evaluated two reference models using the ATIS dataset:
- The first reference is an NLU model which maps ground truth speech transcriptions into intent and entities;
- The second reference model is a Cascaded SLU model which maps speech into transcription then transcription into intent and entities.

These models were also built using transformer based encoders and decoders with exactly the same configuration used in our E2E SLU model. SLU results of these two reference models are shown in Table 2.

### 3.2. Transformer network configuration

The configuration used for the Encoder (E) and Decoder (D) networks in experiments from 1 to 10 are: 8 encoder layers with 2048 units, 4 decoder layers with 2048 units, attention dimension=256, heads=4, ctc_weight=0.5, S2T_weight=0.2, S2I&E_weight=0.2 and transformer-input-layer=conv2d. For experiments 11 and 12 we have increased the number of encoder layers to 12, the decoder layers to 6, the attention dimension to 512 and the number of heads to 8.

| Set | Number | Pre-initialization | | Auxiliary tasks | | Entities (F1 score %) | | Intent (IER %) | |
|---|---|---|---|---|---|---|---|---|---|
| | | Encoder | Decoder | S2T | T2I&E | Filterbank | Wav2vec | Filterbank | Wav2vec |
| 1 | 1 | - | - | - | - | 34.4 | 76.6 | 14.0 | 6.8 |
| | 2 | - | - | - | yes | 53.0 | 83.6 | 13.0 | 4.8 |
| | 3 | - | - | yes | - | 75.1 | 89.1 | 8.5 | 3.9 |
| | 4 | - | - | yes | yes | **87.0** | 89.3 | **5.8** | **3.3** |
| 2 | 5 | ATIS | ATIS | - | - | 88.1 | 90.0 | 3.5 | 3.4 |
| | 6 | ATIS | ATIS | yes | yes | 88.6 | 91.1 | 3.8 | 3.5 |
| | 7 | ATIS | ATIS | yes | - | 90.1 | **91.4** | 3.5 | 3.3 |
| | 8 | ATIS | - | yes | yes | **91.2** | 91.2 | **3.4** | 3.3 |
| 3 | 9 | LibSp100h | - | yes | yes | 90.7 | 89.2 | 3.3 | 3.9 |
| | 10 | LibSp100h | ATIS | yes | yes | 91.1 | **90.0** | 3.3 | **3.0** |
| | 11 | LibSp960h | ATIS | yes | yes | 91.0 | - | **3.1** | - |
| | 12 | LibSp960h | - | yes | yes | **91.2** | - | 3.4 | - |

**Table 1:** Selected experiment using our SLU model. Set 1 - No pre-training, Set 2 - Light pre-training and Set 3 - Enhanced pre-training.

| Ref. models | Entities (F1 score %) | Intent (ER %) |
|---|---|---|
| NLU | 91.6 | 3.2 |
| Cascaded SLU | 88.8 | 3.4 |

**Table 2:** Reference models trained on ATIS dataset.

## 4. RESULTS

From the results of Tables 1 and 2 we highlight the following observations:
- Wav2vec features significantly outperform filterbank features in experiments from Set 1, especially when no auxiliary task are used. They slightly outperform filterbank features in experiments from Set 2, and they slightly underperform compared to filterbank features in experiments from Set 3;
- The highest F1 score for entity prediction is achieved by wav2vec in experiment 7 (Set 2). This F1 score is even higher than the Cascaded SLU reference, and it is very close to the performance of the NLU reference;
- The IER achieved by wav2vec in experiment 4 (Set 1) is either equal or very close to the IER achieved by wav2vec in experiments number 7 and 10 and with the IER achieved by filterbank from experiments 9, 10 and 11;
- Experiments from Set 1 confirm the importance of the auxiliary tasks S2T and T2I&E for the final performance of the proposed E2E SLU model. This result is even stronger for the case of filterbank features;
- From experiments 2 and 4 we can see that auxiliary task T2I&E, which operates only on symbolic domain, mapping word transcriptions into semantic representation (I&E), is also able to improve the performance of the E2E SLU model, mainly in the case of Set 1.

In addition to the experiments from Table 1 we generated two ensemble models [10] using only wav2vec features. The improved results of combining models 3+4 and models 7+8 are shown in Table 3.

| Ensemble models | Entities (F1 score %) | Intent (IER %) |
|---|---|---|
| 3+4 | 90.5 | 3.3 |
| 7+8 | **92.4** | **2.9** |

**Table 3:** Ensemble models using only wav2vec features.

These results clearly demonstrate the efficacy of our proposed approach of using self-supervised features, pre-training and multi-task training for building end-to-end SLU systems.

## 5. CONCLUSION AND FUTURE WORKS

In this work we presented an E2E SLU model based fully on transformer networks which allows easy use of self-supervised pre-trained acoustic features, pre-trained model initialization, and multi-task training. Several experiments using ATIS and LibriSpeech datasets were performed, and we have clearly shown that a simple combination of self-supervised pre-trained acoustic features (like wav2vec) and multi-task learning can significantly improve the performance of E2E SLU models for the cases of light pre-initialization (Set 2) and no pre-initialization at all (Set 1). These results open the possibility of improving E2E SLU models by exploiting huge amount of unlabeled data available for pre-training self-supervised acoustic features.

In future work we intend to (*i*) evaluate other self-supervised acoustic features, such as, wav2vec 2.0, APC, Mockingjay, Tera and audio Alberta; (*ii*) exploit the fine tuning of such self-supervised acoustic features to boost their performance for SLU tasks; and (*iii*) optimize the hyperparameters of the proposed E2E SLU model towards state-of-the-art results.